\documentclass{article}
\usepackage{kr}

\usepackage{times}
\usepackage{soul}
\usepackage{url}
\usepackage[hidelinks]{hyperref}
\usepackage[utf8]{inputenc}
\usepackage[small]{caption}
\usepackage{graphicx}
\usepackage{amsmath}
\usepackage{amsthm}
\usepackage{booktabs}
\usepackage{algorithm}
\usepackage{algorithmic}
\usepackage{mathtools,amssymb}
\usepackage{multirow}
\usepackage{xcolor}
\usepackage{bm}

\usepackage{listings}
\DeclareCaptionFormat{listing}{\rule{\linewidth}{0.08em}{\par}#1#2#3}
\lstdefinestyle{kr}{float=tb,captionpos=t,frame={tb},numbers=left,breaklines=true,columns=fullflexible,keepspaces=true,basicstyle=\ttfamily,xleftmargin=20pt,framexleftmargin=20pt}
\usepackage{tikzsymbols}
\usepackage{tikz}
\usetikzlibrary{arrows,shapes}
\usetikzlibrary{positioning}
\usetikzlibrary{shadows}
\usetikzlibrary{patterns}

\newcommand{\heulingo}[1][]{\ifx\relax#1\relax\textit{heulingo}\else \textit{heulingo}\hspace{0.2em}(#1)\fi}

\newcommand{\clingo}{\textit{clingo}}

\newcommand{\alaspo}{\textit{ALASPO}}
\newcommand{\code}[1]{\lstinline[basicstyle=\ttfamily,breaklines=true]{#1}}
 
\title{Large Neighborhood Prioritized Search for Combinatorial
  Optimization with Answer Set Programming}

\author{Irumi Sugimori$^1$\and
Katsumi Inoue$^2$\and
Hidetomo Nabeshima$^3$\and
Torsten Schaub$^4$\and \\
Takehide Soh$^5$\and
Naoyuki Tamura$^5$\and
Mutsunori Banbara$^1$ \\
\affiliations
$^{1}$Nagoya University, Japan\\$^{2}$National Institute of Informatics, Japan\\$^{3}$University of Yamanashi, Japan\\$^{4}$Universit{\"a}t Potsdam, Germany\\$^{5}$Kobe University, Japan\\}

\begin{document}

\maketitle

\begin{abstract}
 We propose
\emph{Large Neighborhood Prioritized Search} (LNPS)
for solving combinatorial optimization problems
in Answer Set Programming (ASP).
LNPS is a metaheuristic that starts with an initial solution
and then iteratively tries to find better solutions by alternately
destroying and prioritized searching for a current solution.
Due to the variability of neighborhoods,
LNPS allows for flexible search without strongly depending on the
destroy operators.
We present an implementation of LNPS based on ASP.
The resulting {\heulingo} solver demonstrates that LNPS can significantly
enhance the solving performance of ASP for optimization.
Furthermore, we establish the competitiveness of our LNPS approach by
empirically contrasting it to (adaptive) 
large neighborhood search.

 \end{abstract}

\section{Introduction}
\label{sec:introduction}

Systematic search and Stochastic Local Search (SLS) are two major
methods for solving a wide range of combinatorial optimization
problems.
Each method has strengths and weaknesses.
Systematic search can prove the optimality of solutions,
but in general, it does not scale to large instances.
SLS can find near-optimal solutions within a reasonable amount of
time, but it cannot guarantee the optimality of solutions.
Therefore, there has been an increasing interest in the development of
\emph{hybrids} between systematic search and SLS \cite{hoostu15a}.

\emph{Large Neighborhood Search} (LNS) \cite{shaw98a,pisrop19a}
is one of the most studied hybrids.
LNS is an SLS-based metaheuristic that
starts with an initial solution and then iteratively
tries to find better solutions by 
alternately \emph{destroy}ing and \emph{repair}ing a current solution. 
Since the repair operators can be implemented with
systematic solvers,
the LNS heuristic has been shown to be 
highly compatible with 
mixed integer programming \cite{fislod03a,daropa05a} and
constraint programming \cite{shaw98a,debascstta18a,bjflpest19a,bjflpestta20a}.

\emph{Answer Set Programming} (ASP) \cite{lifschitz19a}
is a declarative programming paradigm for
knowledge representation and reasoning.
Due to remarkable improvements in the efficiency of ASP solvers,
ASP has been successfully applied in diverse areas
of artificial intelligence and computer science,
such as
robotics,
computational biology,
product configuration,
decision support,
scheduling,
planning, constraint satisfaction,
model checking,
timetabling,
and many others \cite{ergele16a,aldoma18b,alelge23a,basotainsc13a,bainkaokscsotawa18a}.

The use of LNS with ASP has been recently explored
\cite{eigerumuoest22a}, and soon afterward extended to
\emph{Adaptive Large Neighborhood Search} (Adaptive LNS) \cite{eigerumuoest22b}.
The {\alaspo} solver,
an ASP-based implementation of adaptive LNS,
has demonstrated that LNS can boost
the solving performance of ASP on hard optimization problems \cite{eigerumuoest22b}.
However, LNS strongly depends on the destroy
operators since the undestroyed part is fixed.
In general, it cannot guarantee the optimality of solutions.
It is therefore still particularly challenging to develop a universal
algorithm for ASP which has the advantages of both
systematic search and SLS.

In this paper, we propose
\emph{Large Neighborhood Prioritized Search} (LNPS)
for solving combinatorial optimization problems in ASP.
LNPS is a metaheuristic that starts with an initial solution
and then iteratively tries to find better solutions by alternately
destroying and \emph{prioritized searching} for a current solution.
We present the design and implementation of LNPS based on ASP.
To evaluate the effectiveness of our approach,
we conduct experiments on a benchmark set
used in \cite{eigerumuoest22b}.

The main contributions and results of our paper are
summarized as follows:
\begin{enumerate}
\item We propose Large Neighborhood Prioritized Search (LNPS).
  Since the undestroyed part is not fixed and can be
  prioritized (i.e., \emph{variability}),
  the LNPS heuristic allows for flexible search without
  strongly depending on the destroy operators.
  Moreover, LNPS guarantees the optimality of solutions.

\item We present a design and implementation of LNPS based on ASP.
  In our approach, the LNPS algorithm can be compactly implemented by
  using multi-shot ASP solving and heuristic-driven ASP solving,
  in our case via
  {\clingo}'s Python API \cite{gekakasc17a,karoscwa21a} and
  heuristic statements \cite{gekaotroscwa13a}.

\item The resulting {\heulingo} solver is a tool for
  heuristically-driven answer set optimization.
  {\heulingo} can handle any ASP encodings for
  optimization without any modification.
  All we have to do is to add an LNPS configuration in a declarative way.
  {\heulingo} also supports the traditional LNS heuristic.
  
\item Our empirical analysis considers a challenging benchmark set
  used in \cite{eigerumuoest22b}.
We succeeded in significantly enhancing the solving performance of
  {\clingo} for optimization. 
  Furthermore,
  {\heulingo} demonstrated that the LNPS approach allows us to compete
  with ASP-based adaptive LNS~\cite{eigerumuoest22b}.
\end{enumerate}
Overall, the proposed LNPS can represent a significant contribution to
the state-of-the-art of ASP solving for optimization as well as
hybrids between systematic search and SLS.

 \section{Background}
\label{sec:asp}

In this paper, ASP programs are written in the language of
{\clingo}~\cite{PotasscoUserGuide}.
ASP programs are finite sets of \emph{rules}.
Rules are of the form
\begin{center}
\lstinline[basicstyle=\ttfamily,mathescape=true,breaklines=false]{a$_0$ :- a$_1$,$\dots$,a$_m$,not a$_{m+1}$,$\dots$,not a$_n$.}
\end{center}
Each \lstinline[basicstyle=\ttfamily,mathescape=true,breaklines=true]{a$_i$}
is a propositional \emph{atom}.
An atom \code{a} and its negation \code{not a} are called \emph{literal}.
The left of \code{:-} is a \emph{head}, and the right is a \emph{body}.
The connectives \code{:-}, `\code{,}', and \code{not}
represent if, conjunction, and default negation, respectively.
A period `\code{.}' terminates each rule.
Intuitively, the rule means that
\lstinline[basicstyle=\ttfamily,mathescape=true,breaklines=true]{a$_0$}
must be assigned to true if
\lstinline[basicstyle=\ttfamily,mathescape=true,breaklines=true]{a$_1$}, 
$\dots$, 
\lstinline[basicstyle=\ttfamily,mathescape=true,breaklines=true]{a$_m$}
are true and  
\lstinline[basicstyle=\ttfamily,mathescape=true,breaklines=true]{a$_{m+1}$}, 
$\dots$, 
\lstinline[basicstyle=\ttfamily,mathescape=true,breaklines=true]{a$_n$}
are false.
A rule whose body is empty
(i.e., \lstinline[basicstyle=\ttfamily,mathescape=true,breaklines=true]{a$_0$.})
is called \emph{fact}.
Facts are always true. A rule whose head is empty is called \emph{integrity constraint}:
\begin{center}
\lstinline[basicstyle=\ttfamily,mathescape=true,breaklines=false]{:- a$_1$,$\dots$,a$_m$,not a$_{m+1}$,$\dots$,not a$_n$.}
\end{center}
An integrity constraint represents that the conjunction of literals 
in the body must be false.
Semantically, an ASP program induces a collection of \emph{answer sets}.
Answer sets are distinguished models of the program
based on stable model semantics \cite{gellif88b}.
ASP has some convenient language constructs for
modeling combinatorial (optimization) problems.
A \emph{conditional literal} is of the form 
\lstinline[basicstyle=\ttfamily,mathescape=true,breaklines=true]{$\ell_0$:$\ell_1$,$\dots$,$\ell_m$}.
Each $\ell_i$ is a literal, and 
\lstinline[basicstyle=\ttfamily,mathescape=true,breaklines=true]{$\ell_1$,$\dots$,$\ell_m$}
is called condition like in mathematical set notation.
A \emph{cardinality constraint} of the form
\lstinline[basicstyle=\ttfamily,mathescape=true,breaklines=true]+{$c_1$;$\dots$;$c_n$} = $k$+
represents that exactly $k$ conditional literals among 
\{$c_1$,$\dots$,$c_n$\} must be satisfied.
A weak constraint of the form 
\lstinline[basicstyle=\ttfamily,mathescape=true,breaklines=false]+:$\sim$ $\bm{L}$.[$w$,$\bm{t}$]+
represents preferences in ASP, which is equivalent to
\lstinline[basicstyle=\ttfamily,mathescape=true,breaklines=false]+#minimize {$w$,$\bm{t}$:$\bm{L}$}.+
Here, $w$ is a weight, and $\bm{t}$ and $\bm{L}$ are tuples of terms and literals, respectively.

\begin{lstlisting}[style=kr,basicstyle=\ttfamily\footnotesize,caption={A traditional ASP encoding for TSP solving},label=code:tsp]
{ cycle(X,Y) : edge(X,Y); cycle(X,Y) : edge(Y,X) } = 1 :- vtx(X).
{ cycle(X,Y) : edge(X,Y); cycle(X,Y) : edge(Y,X) } = 1 :- vtx(Y).
reached(1).
reached(Y) :- reached(X), cycle(X,Y).
:- vtx(X), not reached(X).
:~ cycle(X,Y), edgewt(X,Y,C). [C,X,Y]

\end{lstlisting}
\begin{lstlisting}[style=kr,basicstyle=\ttfamily\footnotesize,caption={{\clingo} program activating or deactivating heuristic
  statements on demand for TSP solving},label=code:tsp_heu]
#program heu.
#heuristic cycle(X,Y): heu(cycle(X,Y),W,M). [W,M]

#script(python)
from clingo import Number, Function
def main(ctl):
    ctl.ground([("base",[])])
    ctl.solve()
    a = Function("heu",[Function("cycle",[Number(1),Number(2)]),Number(1),Function("true")])
    ctl.add("ext",[],f"#external {a}.")
    ctl.ground([("ext",[])])
    ctl.ground([("heu",[])])    
    ctl.assign_external(a,True)
    ctl.solve()
    ctl.release_external(a)
    ctl.solve()
#end.



\end{lstlisting}

Multi-shot ASP solving introduces new language constructs:
\code{#program} and \code{#external} statements.
The former statement of the form 
\lstinline[basicstyle=\ttfamily,mathescape=true,breaklines=true]{#program $p(t)$.}
is used to separate an ASP program into several parameterizable subprograms.
The predicate $p$ is a subprogram name and the optional parameter
$t$ is a symbolic constant.
\code{base} is a default subprogram with an empty parameter and includes
rules that are not subject to any \code{#program} statements.
The latter statement of the form 
\lstinline[basicstyle=\ttfamily,mathescape=true,breaklines=false]{#external $a$.} 
represents that the atom $a$ is an \emph{external} atom
whose truth value can be changed later on.
By default, the initial truth value of external atoms is false.
Heuristic-driven ASP solving allows for customizing the
search heuristics of {\clingo} from within ASP programs.
Heuristic information is represented by \code{#heuristic} statements of the form
\lstinline[basicstyle=\ttfamily,mathescape=true,breaklines=false]{#heuristic $a$:$\bm{L}$.[$w$,$m$]}
where $m$ and $w$ are terms representing a heuristic modifier and its
value, respectively.

{\clingo} provides a Python API for controlling ASP's grounding and
solving process.
For illustration, let us consider the well-known Traveling Salesperson Problem (TSP).
A traditional ASP encoding for TSP solving \cite{eigerumuoest22a}
is shown in Listing~\ref{code:tsp}.
The atom \code{cycle(X,Y)} represents that a directed
edge \code{X}~$\rightarrow$~\code{Y} is in a Hamiltonian cycle.
That is, it characterizes a solution.
A {\clingo} program activating or deactivating
heuristic statements on demand for TSP solving is shown in
Listing~\ref{code:tsp_heu}.
This program consists of a subprogram \code{heu} and
an embedded Python script.

Once {\clingo} accepts the program (Listing~\ref{code:tsp_heu})
combined with a TSP instance of fact format and ASP encoding (Listing~\ref{code:tsp}),
a {\clingo} object is created and bound to variable \code{ctl} (cf. Line 6).
The instance and the encoding of Listing~\ref{code:tsp}
are grounded by the \code{ground} function in Line 7.
The \code{solve} function in Line 8 triggers computing stable models
(i.e., solutions of TSP) for the ground program.
That is, the first call of \code{solve} performs a plain TSP solving with {\clingo}.

Next, the external statement of \code{heu(cycle(1,2),1,true)} 
is added to the subprogram \code{ext} by the \code{add} function in Line 10.
This external atom is used to activate or deactivate the heuristic
statement in Line 2.
The external and heuristic statements are grounded 
in Lines 11 and 12, respectively.
Since the truth value of the external atom
is set to true by the \code{assign_external} function in Line 13,
\lstinline[basicstyle=\ttfamily,mathescape=true,breaklines=false]{#heuristic cycle(1,2). [1,true]}
is activated in the second call of \code{solve} in Line 14.
Intuitively, 
this heuristic statement means that
the atom \code{cycle(1,2)} is set to true with higher priority during the search.
More precisely, 
the solver decides first on \code{cycle(1,2)} of level \code{1}
(0 by default for each atom) with a positive sign.
Finally, the heuristic statement is deactivated in the third call of
\code{solve} in Line 16, since 
the truth value of the external atom
is permanently set to false by the \code{release_external} function in Line 15.

 \section{Large Neighborhood Prioritized Search}
\label{sec:lnps}

We consider that the Combinatorial Optimization Problem (COP) is
a minimization problem.
The task of COP is to find a solution $x^{*}$ such that 
$c(x^{*})\leq c(x)\ \forall x\in X$, where
$X$ is the finite set of feasible solutions, and
$c: X\rightarrow \mathbb{R}$ is an objective function that maps from
a solution to its cost.

We propose an SLS-based metaheuristic
called Large Neighborhood Prioritized Search (LNPS)
for solving COPs.
LNPS starts with an initial solution and then iteratively tries to
find better solutions by alternately destroying a current
solution and reconstructing it with prioritized search.
We define the \emph{prioritized search} as a systematic search
for which its branching heuristic can be configured (or customized)
to meet the specific needs of users
based on the priority of 
assignments to
each variable.

\begin{algorithm}[tb]
 \caption{Large Neighborhood Prioritized Search}
 \label{alg:lnps}
 \begin{algorithmic}[1]
  \REQUIRE a feasible solution $x$
  \STATE $x^{*} \gets  x$
  \WHILE {stop criterion is not met}
  \STATE $x^{t} \gets prioritized\mathchar`-search(destroy(x))$
  \IF {$accept(x^{t}, x)$}
  \STATE $x \gets x^{t}$
  \ENDIF
  \IF {$c(x^{t}) < c(x^{*})$}
  \STATE $x^{*} \gets x^{t}$
  \ENDIF
  \ENDWHILE
  \RETURN $x^{*}$
 \end{algorithmic}
\end{algorithm}

The algorithm of LNPS is shown in Algorithm~\ref{alg:lnps}.
Three variables are used in the algorithm.
The variable $x^{*}$ is the best solution obtained during the search.
The variable $x$ is the current solution.
The variable $x^{t}$ is a temporal solution which can be
accepted as the current solution
or discarded.
The $destroy$ operator randomly destroys parts of $x$ and returns
the undestroyed part.
The $prioritized\mathchar`-search$ operator returns a feasible solution,
which is reconstructed from the undestroyed part
by prioritized search.

The best solution $x^{*}$ is initialized in Line 1.
The loop in Lines 2--10 is repeated until
a stop criterion is met.
Typical choices for the stop criterion would include 
the optimality of $x^{*}$, a time-limit, or 
a limit on the number of iterations.
The $destroy$ and $prioritized\mathchar`-search$ operators in Line 3
are alternately applied to find a new solution $x^{t}$.
The new solution is evaluated in Line 4 whether or not 
it becomes the new current solution.
In Line 5, the current solution $x$ is updated if necessary.
There are several ways to implement the \emph{accept} function.
The simplest way is to accept a strictly better solution than the
current solution.
And also,
the new solution is checked in Line 7 whether or not 
it is better than the best known solution.
In Line 8, the current best solution $x^{*}$ is updated if necessary.
Finally, the best solution is returned.

We discuss the main features of LNPS compared with traditional LNS.
In the following figure, each outer rectangle represents a current
solution, in which the dotted part represents the destroyed part.
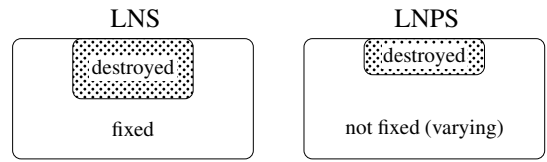
\begin{center}
  \begin{minipage}[t]{0.45\linewidth}\centering
    LNS\\[.3em]
    \scalebox{0.8}{\begin{tikzpicture}
 \draw[rounded corners] (0,0) rectangle (4,2);
 \draw[rounded corners,pattern=crosshatch dots] (1,1) rectangle (3,2);
 \fill[white] (1.27,1.3) rectangle (2.72,1.7);
 \node (d) at (2,1.5) {destroyed};
 \node (f) at (2,0.5) {fixed};
\end{tikzpicture} }
  \end{minipage}
  \begin{minipage}[t]{0.45\linewidth}\centering
    LNPS\\[.3em]
    \scalebox{0.8}{\begin{tikzpicture}
 \draw[rounded corners] (0,0) rectangle (4,2);
 \draw[rounded corners,pattern=crosshatch dots] (1,2) rectangle (3,1.4);
 \fill[white] (1.3,1.9) rectangle (2.7,1.5);
 \node (d) at (2,1.7) {destroyed};
 \node (f) at (2,0.5) {not fixed (varying)};
\end{tikzpicture} }
  \end{minipage}
\end{center}
In LNS \cite{pisrop19a},
the destroy operator, particularly \emph{the percentage of destruction},
plays an essential role since the undestroyed part is fixed.
The percentage of destruction should be sufficiently large such that 
a neighborhood includes better solutions, and be sufficiently small
such that the solver finds one of them.
In addition, LNS cannot guarantee the optimality of solutions in general.

In contrast, 
LNPS can provide flexible search with weakened dependency on the
destroy operators
since the undestroyed part is not fixed (varying) and can be prioritized.
Due to this \emph{variability}, 
the percentage of destruction can be smaller in LNPS.
LNPS can guarantee the optimality of obtained solutions by
appropriately designing a stop criterion of prioritized search.
The easiest way is gradually increasing the time-limit or the
solve-limit on conflicts or restarts.

Furthermore,
the undestroyed part can be configured (or customized)
based on the priority of each variable.
That is, LNPS allows for easy incorporation of domain-specific
heuristics or domain-independent ones into the undestroyed part.
For example, the most simple heuristic would be to keep an initial
solution as much as possible.
This can be achieved by setting the percentage of destruction to zero
and giving high priority to full assignments of decision variables.
Such a heuristic with zero destruction can be useful for quick
\emph{re-scheduling} in real-world applications (e.g., timetabling)
rather than minimal perturbation
with respect to an initial solution~\cite{hawa00,phwaehry17,zigrme11}.
Note that the zero destruction is 
completely useless for LNS since it turns out to be the same solution.

 \section{{\heulingo}: an ASP-based LNPS}
\label{sec:heulingo}

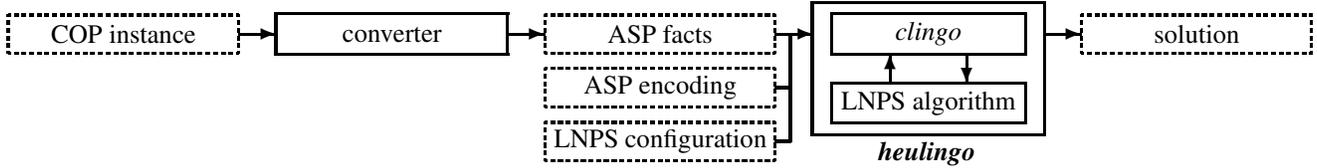
\begin{figure*}[htbp]
  \thicklines
  \setlength{\unitlength}{1.45pt}
  \centering
  ﻿\begin{picture}(340,42)(0,10)
 \put(  0, 35){\dashbox(60,10){\shortstack{COP instance}}}
 \put( 70, 35){\framebox(60,10){converter}}
 \put(140, 35){\dashbox(60,10){\shortstack{ASP facts}}}
 \put(140, 21){\dashbox(60,10){\shortstack{ASP encoding}}}
 \put(140,  7){\dashbox(60,10){\shortstack{LNPS configuration}}}
 \put(210,14){\framebox(60,34){}}
 \put(215, 35){\framebox(50,10){\clingo}}
 \put(215, 17){\framebox(50,10){\shortstack{LNPS algorithm}}}
 \put(280, 35){\dashbox(60,10){\shortstack{solution}}}
 \put( 60, 40){\vector(1,0){10}}
 \put(130, 40){\vector(1,0){10}}
 \put(200, 40){\vector(1,0){10}}
 \put(270, 40){\vector(1,0){10}}
 \put(200, 26){\line(1,0){4}}
 \put(200,  12){\line(1,0){4}}
 \put(204,  12){\line(0,1){28}}
 \put(230,27){\vector(0,1){8}}
 \put(250, 35){\vector(0,-1){8}}
 \put(227,7){\bf\heulingo}
\end{picture}  
   \vskip .5em
  \caption{The architecture of {\heulingo}}
  \label{fig:arch}
\end{figure*}
We develop the {\heulingo} solver which is an LNPS implementation
based on ASP.
The architecture of {\heulingo} is shown in Figure~\ref{fig:arch}.
The {\heulingo} solver accepts a COP instance and an LNPS
configuration in ASP fact format.
In turn, these facts are combined with an ASP encoding for COP solving,
which are afterward solved by the LNPS algorithm
powered by ASP solvers, in our case {\clingo}. ASP facts of LNPS configurations specify
the behavior of the LNPS heuristic, especially for
the destroy and prioritized-search operators.

\begin{algorithm*}[htbp]
  \begin{algorithmic}[1]
  \REQUIRE $P$: problem instance, $E$: ASP encoding, $C$: LNPS configuration
\STATE $ctl.ground([(\text{``\texttt{base}''},\ [])])$ \COMMENT {grounding $P \cup E$}
  \STATE $(ret,\ sol,\ cost) \gets ctl.solve()$
  \IF {$ret = \textrm{OPTIMUM FOUND}$}
  \RETURN $sol$
  \ENDIF
  \STATE $sol\_best,\ cost\_best \gets sol,\ cost$
  \STATE $ctl.ground([(\text{``\texttt{config}''},\ [])])$ \COMMENT {grounding $C$}
  \STATE $lnps\_config \gets get\_lnps\_config()$
  \STATE $rules \gets \text{``''}$
  \FOR {$c$ in $lnps\_config$}
  \STATE $p \gets c.get\_predicate\_name()$
  \STATE $n \gets c.get\_arity()$
  \STATE $atom \gets \text{``}p\texttt{(X}_{1}\texttt{,X}_{2}\texttt{,}\dots\texttt{,X}_{n}\texttt{)}\text{''}$
  \STATE $rules \gets rules + \text{``}\texttt{:- not }atom\texttt{, heuristic(}atom\texttt{,inf,true,t).}\text{''}$ \\
  $\qquad + \text{``}\texttt{:- }atom\texttt{, heuristic(}atom\texttt{,inf,false,t).}\text{''}$ \\
  $\qquad + \text{``}\texttt{\#heuristic }atom\texttt{ : heuristic(}atom\texttt{,W,M,t), W!=inf. [W,M]}\text{''}$
  \ENDFOR
  \STATE $ctl.add(\text{``\texttt{heuristic}''},\ [\text{``\texttt{t}''}],\ rules)$
  \STATE $finished \gets \textrm{False}$
  \STATE $step \gets 0$
  \STATE $prev\_heu\_atoms \gets []$
  \STATE $variability \gets check\_variability(lnps\_config)$
  \WHILE {$finished = \textrm{False}$}
  \STATE $step \gets step + 1$
  \STATE $undestroyed \gets destroy(sol,\ lnps\_config)$
  \STATE $heu\_atoms \gets prioritize(undestroyed,\ lnps\_config,\ step)$
  \STATE $ctl.release\_external(prev\_heu\_atoms)$
  \STATE $statements \gets \text{``''}$
  \FOR {\texttt{heuristic(}$a,w,m,t$\texttt{)} in $heu\_atoms$}
  \STATE $statements \gets statements + \text{``}\texttt{\#external heuristic(}a,w,m,t\texttt{).}\text{''}$
  \ENDFOR
  \STATE $ctl.add(\text{``\texttt{external}''},\ [],\ statements)$
  \STATE $ctl.ground([(\text{``\texttt{external}''},\ [])])$
  \STATE $ctl.ground([(\text{``\texttt{heuristic}''},\ [step])])$
  \FOR {\texttt{heuristic(}$a,w,m,t$\texttt{)} in $heu\_atoms$}
  \STATE $ctl.assign\_external(\texttt{heuristic(}a,w,m,t\texttt{)},\ \textrm{True})$
  \ENDFOR
  \STATE $prev\_heu\_atoms \gets heu\_atoms$
  \STATE $(ret,\ sol\_tmp,\ cost\_tmp) \gets ctl.solve()$
  \IF {$ret = \textrm{OPTIMUM FOUND}$ \AND $variability = \textrm{True}$}
  \STATE $finished \gets \textrm{True}$
  \ENDIF
  \IF {$accept(cost\_tmp,\ cost) = \textrm{True}$}
  \STATE $sol,\ cost \gets sol\_tmp,\ cost\_tmp$
  \ENDIF
  \IF {$cost\_tmp < cost\_best$}
  \STATE $sol\_best,\ cost\_best \gets sol\_tmp,\ cost\_tmp$
  \ENDIF
  \STATE $increase\_solve\_limit()$
  \ENDWHILE
  \RETURN $sol\_best$
 \end{algorithmic}
 \caption{LNPS algorithm with {\clingo}'s multi-shot ASP solving and heuristic statements}
 \label{fig:implementation}
\end{algorithm*}

\textbf{Implementation.}
The pseudo-code of our LNPS algorithm using 
{\clingo}'s multi-shot ASP solving
is shown in Algorithm~\ref{fig:implementation}.
The key idea is utilizing the \code{#heuristic} statement to
implement the prioritized-search operator of LNPS.
The input consists of a problem instance,
an ASP encoding, and an LNPS configuration.
The variables $sol$, $sol\_best$, and $sol\_tmp$ correspond to $x$, $x^*$, and $x^t$
respectively in Algorithm~\ref{alg:lnps}.
The variables $cost$, $cost\_best$, and $cost\_tmp$ represent the
objective values $c(x)$, $c(x^*)$, and $c(x^t)$ respectively.
The variable $ctl$ is a solver object of the \code{Control} class in
{\clingo}'s Python API.
The variable $ret$ is used to store the solving result.

First, in Line 1, 
the instance and the encoding in the \code{base} subprogram are grounded.
The \code{solve} function in Line 2 triggers searching for an
initial solution with a certain stop criterion, and then
the current solution and its cost are initialized.
The algorithm returns the current solution in Line 4
and terminates if it is optimal.
Otherwise, 
the global best solution and its cost are initialized in Line 6.

The LNPS configuration (see below for details) is grounded and parsed
in Lines 7--15.
The heuristic statements of the form
\lstinline[basicstyle=\ttfamily,mathescape=true,breaklines=true]+#heuristic $p(X_{1},\ldots,X_{n})$:heuristic($p(X_{1},\ldots,X_{n})$,W,M,t), W!=inf. [W,M]+
are added to the \code{heuristic} subprogram in Line 16.
The external atom 
\lstinline[basicstyle=\ttfamily,mathescape=true,breaklines=true]{heuristic($p(X_{1},\ldots,X_{n})$,W,M,t)}
in the body 
is used to activate or deactivate the heuristic statements on demand.
We refer to it as \emph{heuristic atom}.
In addition, two kinds of integrity constraints in Line 14 are added
in a similar way to support traditional LNS.

Next, the loop in Lines 21--48 is repeated until
a stop criterion is met.
The variable $finished$,
initialized to False in Line 17,
is a flag for whether iterations should be finished or not.
The $check\_variability$ function in Line 20 checks
the variability of the LNPS configuration.
That is, the variable $variability$ is set to False if
the undestroyed part may be fixed for LNS, otherwise True.

In each iteration, 
the algorithm invokes the $destroy$ function in Line 23 to destroy
parts of the current solution, according to the configuration.
In turn, the $prioritize$ function in Line 24 is invoked to 
generate new heuristic atoms for the undestroyed part.
The heuristic statements in the previous iteration are 
deactivated in Line 25.
The external statements for new heuristic atoms
are generated and added to the \code{external} subprogram
in Lines 27--30.

We are now ready to try to find a better feasible solution.
The external and heuristic statements are grounded 
in Lines 31 and 32, respectively.
The heuristic statements are activated in Lines 33--35 by setting the
truth value of their heuristic atoms to true.
The \code{solve} function in Line 37 triggers heuristically searching
for a new solution,
and then the temporal solution and its cost are updated.
Note that 
the stop criterion of \code{solve} in Lines 2 and 37 can be
separately specified using {\clingo}'s options:
the time-limit in seconds (\code{--time-limit})  or
the solve-limit on conflicts or restarts (\code{--solve-limit}).
We adopt the latter solve-limit in 
the current implementation of {\heulingo}.

In Line 38, the termination criterion is checked.
The variable $finished$ is set to True in Line 39
if the new solution is optimal and
the $variability$ flag is True.
When the variability flag is False (i.e., LNS),
the algorithm cannot terminate even if the new solution is optimal
since the undestroyed part is fixed.
The new solution is evaluated in Line 41 whether it can become
the current solution or should be rejected.
{\heulingo} accepts only improving solution by default, 
but the acceptance criterion can be customized by a {\heulingo}'s option.
And also, the new solution is checked in Line 44 
whether or not it is better than the best known solution.
In Line 45, the current best solution is updated if necessary.
At the end of the loop in Line 47, 
the value of {\clingo}'s solve-limit is increased
by the function $increase\_solve\_limit$
to guarantee the optimality of solutions.
Finally, the algorithm returns the best solution in Line 49.

\textbf{ASP fact format of LNPS configurations.}
We introduce three different kinds of predicates
to specify configurations of the LNPS heuristic
in the \code{config} subprogram.
The predicate \code{_lnps_project/2} is used to define what
subset of the atoms belonging to an answer set is subject to LNPS.
We refer to the atoms via \code{_lnps_project/2} as \emph{projected atoms}.
In general, the projected atoms characterize a solution. 
The predicate \code{_lnps_destroy/4} is used to define
what part of the projected atoms is destroyed and by what percentage.
The predicate \code{_lnps_prioritize/4} is used to define 
how the projected atoms in the undestroyed part are prioritized
(or fixed).

For illustration, let us consider a configuration of the LNPS heuristic
for the TSP encoding in Listing~\ref{code:tsp}.
For this, the \emph{random destruction} would be one of the 
most simple configurations, which is shown in Listing~\ref{code:random_N}.
Intuitively, the configuration represents a heuristic that randomly
destroys a current solution, and the undestroyed part is not
fixed but is kept as much as possible in each iteration.
The atom \code{_lnps_project(cycle,2)} characterizes a solution
as the atoms of \code{cycle/2} belonging to an answer set.
The atom \code{_lnps_destroy(cycle,2,3,p(n))} means that
the $destroy$ function in Line 23 of Algorithm~\ref{fig:implementation}
randomly destroys \code{n}\% of the projected atoms of \code{cycle/2}.
Note that the third argument \code{3}=$(11)_{2}$ represents that all
possible two arguments (\code{X},\code{Y}) such that \code{cycle(X,Y)} holds
are subject to destruction.
The atom \code{_lnps_prioritize(cycle,2,1,true)} means that a statement
\lstinline[basicstyle=\ttfamily,mathescape=true,breaklines=true]+#heuristic cycle(X,Y) : heuristic(cycle(X,Y),W,M,t), W != inf. [W,M]+
is added to the \code{heuristic} subprogram in Line 16 of
Algorithm~\ref{fig:implementation}.
The traditional LNS heuristic of fixing the undestroyed part can be
done by replacing a fact in Line 4 of Listing~\ref{code:random_N} with 
\lstinline[basicstyle=\ttfamily,mathescape=true,breaklines=true]+_lnps_prioritize(cycle,2,inf,true).+

\begin{lstlisting}[style=kr,basicstyle=\ttfamily\footnotesize,caption={A simple LNPS heuristic for TSP solving},label=code:random_N]
#program config.
_lnps_project(cycle,2).
_lnps_destroy(cycle,2,3,p(n)).
_lnps_prioritize(cycle,2,1,true).
\end{lstlisting}

\textbf{The main features of {\heulingo}}.
The {\heulingo} solver can be considered as a tool for heuristically-driven
answer set optimization.
In addition to the \emph{variability} and \emph{optimality} from LNPS,
{\heulingo} has the following features.
\begin{itemize}
\item \emph{Expressiveness}:
  {\heulingo} relies on ASP's expressive language that is well suited
  for modeling combinatorial optimization problems.
\item\emph{Implementation}:
  The LNPS algorithm can be compactly implemented using
  {\clingo}'s  multi-shot ASP solving and heuristic statements,
  as can be seen in Algorithm~\ref{fig:implementation}.
\item\emph{Domain heuristics}:
  {\heulingo} allows for easy incorporation of 
  domain heuristics in a declarative way, such as 
  the random destruction in Listing~\ref{code:random_N}.
  More sophisticated domain-specific heuristics can also be
  incorporated.
\item\emph{Usability and Compatibility}: 
  {\heulingo} can deal with any ASP encoding for
  optimization without any modification.
  All we have to do is to add an LNPS configuration.
  {\heulingo} also supports the traditional LNS heuristic.
\end{itemize}
For \emph{efficiency},
the question is whether the {\heulingo} approach matches the
performance of the (adaptive) LNS heuristic.
We empirically address this question 
in the next section.

 \section{Experiments}
\label{sec:experiments}

\begin{table*}[tb]
 \centering 
 \begin{tabular}{c|r|rrr|rrr|rrr}
\toprule
\multirow{2}{*}{Instance} & \multicolumn{1}{c|}{\multirow{2}{*}{\clingo}} & \multicolumn{3}{c|}{{\heulingo[LNS]}} & \multicolumn{3}{c|}{{\heulingo[LNPS]}} & \multicolumn{3}{c}{\alaspo}\\
& & \multicolumn{1}{c}{avg.} & \multicolumn{1}{c}{min.} & \multicolumn{1}{c|}{max.} & \multicolumn{1}{c}{avg.} & \multicolumn{1}{c}{min.} & \multicolumn{1}{c|}{max.} & \multicolumn{1}{c}{avg.} & \multicolumn{1}{c}{min.} & \multicolumn{1}{c}{max.} \\
\midrule
dom\_rand\_70\_300\_1155482584\_3 & 591 & 438.7 & 427 & 454 & {\textbf{386.3}} & 383 & 390 & 424.3 & 397 & 444 \\
rand\_70\_300\_1155482584\_0 & 552 & 371.7 & 351 & 393 & {\textbf{326.3}} & 320 & 333 & 367.7 & 349 & 384 \\
rand\_70\_300\_1155482584\_11 & 606 & 447.0 & 436 & 454 & {\textbf{386.3}} & 381 & 392 & 447.0 & 433 & 466 \\
rand\_70\_300\_1155482584\_12 & 540 & 386.3 & 364 & 406 & {\textbf{344.7}} & 341 & 349 & 380.7 & 371 & 386 \\
rand\_70\_300\_1155482584\_14 & 567 & 393.7 & 388 & 404 & {\textbf{357.7}} & 355 & 359 & 397.0 & 382 & 409 \\
rand\_70\_300\_1155482584\_3 & 575 & 444.7 & 428 & 458 & {\textbf{408.7}} & 398 & 419 & 450.0 & 445 & 459 \\
rand\_70\_300\_1155482584\_4 & 649 & 476.7 & 464 & 483 & {\textbf{423.0}} & 419 & 428 & 475.7 & 470 & 479 \\
rand\_70\_300\_1155482584\_5 & 601 & 420.0 & 397 & 449 & {\textbf{367.3}} & 361 & 374 & 396.3 & 393 & 401 \\
rand\_70\_300\_1155482584\_7 & 604 & 435.0 & 429 & 446 & {\textbf{406.3}} & 405 & 407 & 442.0 & 428 & 462 \\
rand\_70\_300\_1155482584\_8 & 553 & 441.7 & 426 & 461 & {\textbf{387.0}} & 385 & 389 & 427.0 & 412 & 441 \\
rand\_70\_300\_1155482584\_9 & 546 & 414.3 & 391 & 427 & {\textbf{368.3}} & 365 & 372 & 403.3 & 402 & 405 \\
rand\_80\_340\_1159656267\_0 & 714 & 464.7 & 446 & 492 & {\textbf{410.7}} & 410 & 411 & 479.0 & 476 & 484 \\
rand\_80\_340\_1159656267\_10 & 654 & 494.0 & 480 & 503 & {\textbf{441.3}} & 438 & 445 & 499.7 & 495 & 507 \\
rand\_80\_340\_1159656267\_11 & 731 & 528.7 & 509 & 539 & {\textbf{464.0}} & 458 & 475 & 520.7 & 497 & 534 \\
rand\_80\_340\_1159656267\_13 & 686 & 467.7 & 437 & 487 & {\textbf{431.3}} & 426 & 440 & 471.3 & 466 & 477 \\
rand\_80\_340\_1159656267\_15 & 720 & 492.7 & 484 & 499 & {\textbf{439.3}} & 435 & 446 & 478.0 & 471 & 488 \\
rand\_80\_340\_1159656267\_16 & 667 & 546.7 & 525 & 559 & {\textbf{496.3}} & 492 & 499 & 558.7 & 551 & 571 \\
rand\_80\_340\_1159656267\_17 & 737 & 501.3 & 492 & 509 & {\textbf{449.0}} & 443 & 457 & 472.3 & 461 & 479 \\
rand\_80\_340\_1159656267\_18 & 674 & 484.7 & 466 & 510 & {\textbf{418.7}} & 417 & 420 & 488.0 & 477 & 506 \\
rand\_80\_340\_1159656267\_4 & 590 & 471.7 & 442 & 511 & {\textbf{418.3}} & 413 & 421 & 462.3 & 460 & 466 \\
\midrule
Average rate & 1.000 & 0.728 &  &  & {\textbf{0.649}} &  &  & 0.721 &  &  \\
\bottomrule
\end{tabular}
  \caption{Comparison results on traveling salesperson problem}
 \label{tab:tsp}
\end{table*}

To evaluate our approach, we carry out experiments on a challenging
benchmark set and ASP encodings used in \cite{eigerumuoest22b}.
The benchmark set consists of
Traveling Salesperson Problem (TSP),
Social Golfer Problem (SGP), 
Sudoku Puzzle Generation (SPG), 
Weighted Strategic Companies (WSC), 
and Shift Design (SD).
The ASP encodings include an encoding for TSP solving in
Listing~\ref{code:tsp}.

We compare {\heulingo[LNPS]} and {\heulingo[LNS]} with
{\clingo} and {\alaspo}.
Here, {\heulingo[$X$]} indicates that {\heulingo} uses heuristic $X$.
\begin{itemize}
\item We run {\clingo}-5.6.2~\footnote{\url{https://potassco.org/clingo/}}
  with the default configuration unless otherwise noted.
\item We execute {\heulingo} in 3 runs for each instance using
  the random destruction with different percentages.
\item We execute {\alaspo}~\footnote{\url{http://www.kr.tuwien.ac.at/research/projects/bai/kr22.zip}}
  in 3 runs for each instance 
  with the best portfolio~\footnote{We use dynamic.json for TSP and SPG,
    roulette\_alpha0.4.json for SGP,
    roulette\_alpha0.8.json for WSC, and
    uniform.json for SD.}
  presented in \cite{eigerumuoest22b}.
\end{itemize}
{\alaspo} is an ASP-based implementation of adaptive LNS.
{\alaspo} selects in each iteration
a potentially more effective destroy operator
from a pre-defined portfolio.
The portfolio consists of a selection strategy,
multiple destroy operators, and their percentages of destruction.
{\alaspo} implements three selection strategies:
self-adaptive roulette-wheel strategy, 
uniform roulette-wheel strategy,
and dynamic strategy.
{\alaspo} provides two destroy operators:
random-atoms and random-constants.
The random-atoms operator is the same as the random destruction as
explained in Section~\ref{sec:heulingo}.
The random-constants operator randomly selects a sample from all
constants of the atoms via \#show statements, and
destroys all atoms containing any selected constants.

We use Python 3.9.18 to run {\heulingo} and {\alaspo}.
We run TSP, SGP, and SPG on Mac OS Apple M1 Ultra (20-core CPU and 128GB memory),
WSC on Mac OS Intel Xeon W (12-core CPU and 96GB memory), and
SD on Mac OS Apple M1 (8-core CPU and 16GB memory).

\textbf{Traveling salesperson problem}
is a well-known optimization problem.
The task is to find a Hamiltonian cycle of 
minimum accumulated edge costs.
The time-limit is 300s for each instance.
The solve-limit of {\heulingo} is set to 
1,210,000 and 800,000 conflicts
for finding an initial solution and each iteration, respectively.
We use three different percentages of the random destruction:
\{1\%, 3\%, 5\%\} for {\heulingo[LNPS]} and \{28\%, 30\%, 32\%\} for {\heulingo[LNS]}.

Comparison results of obtained bounds are shown in Table~\ref{tab:tsp}.
The column shows in order
the instance names,
the obtained bounds of {\clingo}, 
and the average, minimum, and maximum of obtained bounds in 3 runs of 
{\heulingo[LNS]}, {\heulingo[LNPS]}, and {\alaspo}.
The best average in each row is highlighted in bold.
The bottom shows the average rates to the bounds obtained by {\clingo}.
{\heulingo[LNPS]} is able to find the best bounds on average for all 20 instances.
{\heulingo[LNPS]} succeeds in improving the bounds of {\clingo} by 35.1\% on average.

\begin{table*}[tb]
 \centering 
 \begin{tabular}{c|r|rrr|rrr|rrr}
\toprule
\multirow{2}{*}{\#weeks ($w$)} & \multicolumn{1}{c|}{\multirow{2}{*}{\clingo}} & \multicolumn{3}{c|}{{\heulingo[LNS]}} & \multicolumn{3}{c|}{{\heulingo[LNPS]}} & \multicolumn{3}{c}{\alaspo}\\
& & \multicolumn{1}{c}{avg.} & \multicolumn{1}{c}{min.} & \multicolumn{1}{c|}{max.} & \multicolumn{1}{c}{avg.} & \multicolumn{1}{c}{min.} & \multicolumn{1}{c|}{max.} & \multicolumn{1}{c}{avg.} & \multicolumn{1}{c}{min.} & \multicolumn{1}{c}{max.} \\
\midrule
8 & 3 & 2.0 & 2 & 2 & {\textbf{1.7}} & 1 & 2 & 2.0 & 2 & 2 \\
9 & 7 & {\textbf{4.3}} & 4 & 5 & 4.7 & 4 & 5 & 5.0 & 5 & 5 \\
10 & 10 & {\textbf{7.0}} & 7 & 7 & 7.7 & 7 & 8 & 7.7 & 7 & 8 \\
11 & 11 & 10.0 & 10 & 10 & {\textbf{9.7}} & 9 & 10 & 10.3 & 10 & 11 \\
12 & 15 & {\textbf{13.0}} & 13 & 13 & {\textbf{13.0}} & 13 & 13 & 13.7 & 13 & 14 \\
\midrule
Average rate & 1.00000 & 0.75134 &  &  & {\textbf{0.75132}} &  &  & 0.80013 &  &  \\
\bottomrule
\end{tabular}
  \caption{Comparison results on social golfer problem}
 \label{tab:golfer}
\end{table*}
\begin{table*}[tb]
 \centering 
 \begin{tabular}{c|r|rrr|rrr|rrr}
\toprule
\multirow{2}{*}{Size} & \multicolumn{1}{c|}{\multirow{2}{*}{\clingo}} & \multicolumn{3}{c|}{{\heulingo[LNS]}} & \multicolumn{3}{c|}{{\heulingo[LNPS]}} & \multicolumn{3}{c}{\alaspo}\\
& & \multicolumn{1}{c}{avg.} & \multicolumn{1}{c}{min.} & \multicolumn{1}{c|}{max.} & \multicolumn{1}{c}{avg.} & \multicolumn{1}{c}{min.} & \multicolumn{1}{c|}{max.} & \multicolumn{1}{c}{avg.} & \multicolumn{1}{c}{min.} & \multicolumn{1}{c}{max.} \\
\midrule
$9\times 9$ & 21 & 20.0 & 19 & 21 & {\textbf{19.3}} & 19 & 20 & 20.0 & 20 & 20 \\
$16\times 16$ & 232 & 95.3 & 93 & 98 & {\textbf{93.0}} & 90 & 95 & 96.7 & 95 & 99 \\
\midrule
Average rate & 1.000 & 0.682 &  &  & {\textbf{0.660}} &  &  & 0.685 &  &  \\
\bottomrule
\end{tabular}
  \caption{Comparison results on sudoku puzzle generation}
 \label{tab:sudoku}
\end{table*}

\begin{table*}[tb]
 \centering 
 \tabcolsep=0.5em
 \begin{tabular}{c|r|rrr|rrr|rrr}
\toprule
\multirow{2}{*}{Instance} & \multicolumn{1}{c|}{\multirow{2}{*}{\clingo}} & \multicolumn{3}{c|}{{\heulingo[LNS]}} & \multicolumn{3}{c|}{{\heulingo[LNPS]}} & \multicolumn{3}{c}{\alaspo}\\
& & \multicolumn{1}{c}{avg.} & \multicolumn{1}{c}{min.} & \multicolumn{1}{c|}{max.} & \multicolumn{1}{c}{avg.} & \multicolumn{1}{c}{min.} & \multicolumn{1}{c|}{max.} & \multicolumn{1}{c}{avg.} & \multicolumn{1}{c}{min.} & \multicolumn{1}{c}{max.} \\
\midrule
wstratcomp\_001 & 198988 & {\textbf{194247.3}} & 192783 & 195411 & 195288.3 & 194517 & 196179 & 204601.7 & 200494 & 208733 \\
wstratcomp\_006 & 81819 & 80058.0 & 79390 & 80392 & 78144.3 & 77420 & 78928 & {\textbf{75128.7}} & 74544 & 75654 \\
wstratcomp\_015 & 163906 & {\textbf{160918.0}} & 160718 & 161288 & 161968.3 & 161828 & 162150 & 198342.3 & 190565 & 212559 \\
wstratcomp\_018 & 129784 & 135377.3 & 135204 & 135724 & 131159.0 & 128835 & 133126 & {\textbf{118672.7}} & 114983 & 121040 \\
wstratcomp\_019 & 94978 & 96533.0 & 96533 & 96533 & 96194.7 & 95518 & 96533 & {\textbf{91169.7}} & 90602 & 91664 \\
wstratcomp\_030 & 182200 & 180349.7 & 179806 & 181285 & {\textbf{179982.0}} & 179535 & 180563 & 210109.7 & 208706 & 212666 \\
wstratcomp\_033 & {\textbf{193568}} & 196136.3 & 195765 & 196330 & 194886.3 & 193875 & 196009 & 220024.3 & 218604 & 222432 \\
wstratcomp\_042 & 133273 & 134648.0 & 134164 & 135235 & 130933.0 & 129125 & 132450 & {\textbf{117355.3}} & 115786 & 119145 \\
wstratcomp\_050 & 166498 & {\textbf{166067.0}} & 165159 & 166859 & 166249.0 & 165667 & 167011 & 190745.3 & 188446 & 193819 \\
wstratcomp\_051 & 69582 & 60815.7 & 60196 & 61432 & {\textbf{59637.0}} & 58882 & 60077 & 63122.0 & 62015 & 64043 \\
wstratcomp\_052 & 70868 & 62591.3 & 62508 & 62677 & {\textbf{62044.3}} & 61927 & 62188 & 63597.7 & 63370 & 63948 \\
wstratcomp\_053 & 72848 & 63239.0 & 62312 & 63749 & {\textbf{63175.0}} & 63082 & 63276 & 64441.7 & 64136 & 64955 \\
wstratcomp\_054 & 75352 & 63511.0 & 62482 & 64091 & {\textbf{62294.7}} & 62082 & 62492 & 63903.7 & 61825 & 66365 \\
wstratcomp\_055 & 77469 & 67033.7 & 66686 & 67277 & {\textbf{66889.3}} & 66823 & 67020 & 68190.7 & 67616 & 69190 \\
wstratcomp\_056 & 68919 & 61420.0 & 60967 & 62309 & {\textbf{61161.0}} & 60798 & 61616 & 66436.3 & 65371 & 67025 \\
wstratcomp\_057 & 67836 & 63465.7 & 63259 & 63806 & {\textbf{62630.7}} & 62519 & 62762 & 66218.3 & 66013 & 66570 \\
wstratcomp\_058 & 73174 & 63477.0 & 62912 & 64583 & {\textbf{61092.7}} & 61003 & 61256 & 65160.0 & 64094 & 66142 \\
wstratcomp\_059 & 71986 & 62733.0 & 62641 & 62828 & {\textbf{61061.3}} & 60615 & 61664 & 62271.0 & 61454 & 62885 \\
wstratcomp\_060 & 75302 & 66103.0 & 65476 & 66649 & {\textbf{64355.0}} & 64289 & 64399 & 68726.3 & 68383 & 69398 \\
wstratcomp\_061 & 70918 & 63846.7 & 63640 & 64116 & {\textbf{63805.3}} & 63561 & 64084 & 65452.7 & 65173 & 65842 \\
\midrule
Average rate & 1.000 & 0.934 &  &  & {\textbf{0.923}} &  &  & 0.965 &  &  \\
\bottomrule
\end{tabular}
  \caption{Comparison results on weighted strategic companies}
 \label{tab:wsc}
\end{table*}

\textbf{Social golfer problem}
is a combinatorial optimization problem
whose goal is to schedule $g$ groups of $p$ players for $w$ weeks
such that no two golfers play together more than once.
We consider instances with $g=8$, $p=4$, and $8\leq w\leq 12$.
The time-limit is 1,800s for each $w$.
The solve-limit of {\heulingo} is set to 
500,000 conflicts for finding an initial solution and each iteration.
We use three different percentages of the random destruction:
\{55\%, 60\%, 65\%\} for {\heulingo[LNPS]} and \{60\%, 65\%, 70\%\} for {\heulingo[LNS]}.
In addition,
we use {\heulingo}'s option to limit the deterioration of objective values during the search.
More precisely, this option enforces that every time a current
solution and its cost are updated, the cost minus one is set
to the initial bound for the objective function in the next iteration.

Comparison results of obtained bounds are shown in Table~\ref{tab:golfer}.
{\heulingo} is able to find the best bounds on average for all $8\leq w\leq 12$.
{\heulingo[LNPS]} provides the same or better bounds on minimum for
all $w$ than {\heulingo[LNS]}.
{\heulingo[LNPS]} succeeds in improving the bounds of {\clingo} by 24.868\% on average.
{\heulingo} performs slightly better on average than {\alaspo}.

\textbf{Sudoku puzzle generation}
is an optimization problem 
whose goal is to find an $N\times N$ sudoku puzzle
with a minimal number of hints.
We consider two sizes of $N=9$ and $N=16$.
A disjunctive ASP encoding~\cite{eigerumuoest22a} we used
takes advantage of
a \emph{saturation} technique~\cite{eitgot95a}.
The time-limit is 600s for each size $N$.
We use {\clingo}'s options \code{--configuration=many} and  \code{-t4} as
with \cite{eigerumuoest22b} for all solvers.
The solve-limit of {\heulingo} is set to 
300,000 and 6,000 conflicts for finding an initial solution and each
iteration, respectively.
We use three different percentages of the random destruction:
\{12\%, 14\%, 16\%\} for {\heulingo[LNPS]} and \{18\%, 20\%, 22\%\}
for {\heulingo[LNS]}.

Comparison results of the obtained number of hints 
are shown in Table~\ref{tab:sudoku}.
{\heulingo[LNPS]} is able to find the best bounds on average for both
of the two sizes.
{\heulingo[LNPS]} succeeds in improving the bounds of {\clingo} by
34\% on average.
{\heulingo} provides a near-optimal solution of 19 hints
for $N=9$, compared with the known minimal hint of 17.

\textbf{Weighted strategic companies}~\cite{eigerumuoest22a} 
is an optimization variant of the $\Sigma_{2}^{P}$-hard
strategic companies problem~\cite{caeigo97a}.
The task is to find strategic sets such that the sum of weights of
strategic companies is minimized.
The time-limit is 3,600s for each instance.
We use {\clingo}'s option \code{--opt-strategy=usc,15}
for finding an initial solution in {\heulingo} and {\alaspo}.~\footnote{We use this option because it provides better bounds than the
  default configuration in our preliminary experiments.} 
The solve-limit of {\heulingo} is set to 
30,000,000 and 60,000 conflicts for finding an initial solution and
each iteration respectively on the first 9 instances in
Table~\ref{tab:wsc}, and to 
1,000,000 and 40,000 conflicts on the next 11 instances.
We use three different percentages of the random destruction:
\{4\%, 7\%, 10\%\} for {\heulingo[LNPS]} and \{16\%, 19\%, 22\%\} for {\heulingo[LNS]}.
 
Comparison results of obtained bounds are shown in Table~\ref{tab:wsc}.
{\heulingo[LNPS]} is able to find the best bounds on average for
12 instances, compared with 
1 of {\clingo},
3 of {\heulingo[LNS]}, and
4 of {\alaspo}.
Although {\clingo} performs well on WSC,
{\heulingo[LNPS]} succeeds in improving the bounds of {\clingo} by 7.7\% on average.
{\heulingo} performs slightly better on average than {\alaspo}.

\begin{table*}[tb]
 \centering 
 \begin{tabular}{c|r|rr|rr|rr}
\toprule
\multirow{2}{*}{Instance} & \multicolumn{1}{c|}{\multirow{2}{*}{\clingo}} & \multicolumn{2}{c|}{{\heulingo[LNS]}} & \multicolumn{2}{c|}{{\heulingo[LNPS]}} & \multicolumn{2}{c}{\alaspo} \\
& & \multicolumn{1}{c}{min.} & \multicolumn{1}{c|}{max.} & \multicolumn{1}{c}{min.} & \multicolumn{1}{c|}{max.} & \multicolumn{1}{c}{min.} & \multicolumn{1}{c}{max.} \\
\midrule
4\_30m & (0, 427, 50) & (0, 310, 19) & (0, 311, 22) & (0, 353, 45) & (0, 375, 44) & {\textbf{(0, 310, 14)}} & (0, 310, 18) \\
6\_15m & (0, 266, 46) & {\textbf{(0, 175, 20)}} & (0, 176, 24) & (0, 187, 33) & (0, 202, 35) & (0, 176, 31) & (0, 189, 38) \\
11\_30m & (0, 821, 66) & {\textbf{(0, 643, 46)}} & (0, 694, 61) & (0, 709, 63) & (0, 719, 64) & (0, 645, 61) & (0, 685, 60) \\
20\_30m & (0, 1035, 60) & {\textbf{(0, 900, 49)}} & (0, 951, 62) & (0, 944, 65) & (0, 952, 64) & (0, 906, 55) & (0, 926, 58) \\
26\_30m & (0, 1062, 78) & {\textbf{(0, 771, 41)}} & (0, 771, 48) & (0, 991, 78) & (0, 1004, 82) & (0, 784, 68) & (0, 898, 78) \\
27\_60m & (0, 393, 25) & (0, 362, 16) & (0, 362, 18) & (0, 387, 26) & (0, 393, 25) & {\textbf{(0, 362, 15)}} & (0, 362, 17) \\
29\_30m & (0, 528, 59) & {\textbf{(0, 447, 38)}} & (0, 450, 45) & (0, 459, 57) & (0, 460, 59) & (0, 447, 42) & (0, 451, 48) \\
2\_30m & (0, 456, 52) & (0, 388, 22) & (0, 388, 28) & (0, 394, 45) & (0, 398, 54) & {\textbf{(0, 388, 14)}} & (0, 388, 16) \\
\bottomrule
\end{tabular}
  \caption{Comparison results on shift design}
 \label{tab:sd}
\end{table*}

\textbf{Shift design} is an employee scheduling problem.
The task is to find staff schedules considering
the minimization of the number of shifts
and both over- and understaffing.
The ASP encoding~\cite{abgemuscwo15c} we used is based on
lexicographic optimization of three objective functions.
The time-limit is 3,600s for each instance.
We use {\clingo}'s options
\code{--configuration=handy} and \code{--opt-strategy=usc,3}
as with \cite{eigerumuoest22b}
for all solvers.
The solve-limit of {\heulingo} is set to 
900,000 and 40,000 conflicts for finding an initial solution and each
iteration, respectively.
We use three different percentages of the random destruction:
\{10\%, 20\%, 30\%\} for {\heulingo[LNPS]} and 
\{25\%, 35\%, 45\%\} for {\heulingo[LNS]}.
We use {\heulingo}'s option to limit the deterioration of
objective values during the search, as with SGP.

Comparison results of obtained bounds are shown in Table~\ref{tab:sd}.
{\heulingo[LNS]} is able to find the best bounds for
5 among 8 instances. Although {\heulingo[LNPS]} can find better bounds for all instances than {\clingo},
it does not match the performance of {\heulingo[LNS]} and {\alaspo}.

\textbf{Summary and discussion.}
{\heulingo[LNPS]} was able to find the best bounds on average for 
37 among all 55 instances (67\% in a total).
{\heulingo[LNPS]} succeeded in improving the bounds of {\clingo} by
35.1\% for TSP,
24.8\% for SGP,
34.0\% for SPG, and
7.7\% for WSC.
On the other hand, however,
{\heulingo[LNPS]} met the difficulty of decreasing the bounds
sufficiently on shift design.
To resolve this issue, by our results,
it would be effective to extend {\heulingo} for
adaptive LNPS combining the LNS and LNPS heuristics.

In general, it can be a hard and time-consuming task to find the
best configuration for LNPS.
The configurations of {\heulingo} were obtained in our
preliminary experiments.
We tested, for each benchmark problem, some percentages of
destruction less than the ones used in \cite{eigerumuoest22a},
taking the variability of LNPS into account.
And then, we selected the best one and two values below and above it.

We observed that 
{\clingo} quickly falls into saturated solutions for many instances, 
in the sense that it has the difficulty of decreasing the bounds
sufficiently.
Thus, we roughly estimated the number of conflicts on which 
{\clingo} is stuck at saturated solutions, and then used it for the
solve-limit of {\heulingo}.
We plan to extend the functionality of {\clingo}'s API
to measure whether or not {\clingo} gets stuck during the search,
and it will be helpful to develop not only 
adaptive LNPS but also other metaheuristics in ASP.

 \section{Related Work}
\label{sec:related_work}

New techniques for optimization in ASP have been continually
developed, such as
core-guided optimization~\cite{alvdod16a,alvdod17a,ankamasc12a} and
complex preference handling~\cite{brewka04c,brderosc15a,gekasc11b}.
Multi-shot ASP solving~\cite{gekakasc17a,karoscwa21a} and 
heuristic-driven ASP solving~\cite{dogalemurisc16a,gekaotroscwa13a,gerysc15a}
can be quite useful for implementing
widely different SLS-based metaheuristics in ASP.

Besides work on LNS in ASP~\cite{eigerumuoest22a,eigerumuoest22b},
LNS has been used in combination with
MaxSAT~\cite{demmus17},
mixed integer programming~\cite{fislod03a,daropa05a}, and 
constraint programming~\cite{shaw98a,debascstta18a,bjflpest19a,bjflpestta20a}.
The use of SLS for SAT,
considering variable dependencies~\cite{kaubar07},
was studied in~\cite{bejast11}.

On the traditional LNS heuristic,
starting with~\cite{shaw98a},
a great deal of research has been done~\cite{pisrop19a}.
LNS and its extensions have been so far successfully
applied in the areas of routing and scheduling problems, including
vehicle routing problems~\cite{shaw98a},
pickup and delivery problem with time windows~\cite{ropi06},
cumulative job shop scheduling problem~\cite{golanu05a},
and
technician and task scheduling problem~\cite{colaparo10a}.

More recently, 
multi-agent path finding~\cite{lichhastko21a,phhudiko24a},
timetabling~\cite{kihasc17a,demmus17}, and
test laboratory scheduling~\cite{gemimu21a}
are well explored.
By our results, it would be interesting to explore
whether the variability feature of LNPS could be 
effective for these problems.

 \section{Conclusion}
\label{sec:conclusion}

We proposed
Large Neighborhood Prioritized Search (LNPS)
for solving combinatorial optimization problems.
We presented an implementation design of LNPS based on Answer Set
Programming (ASP).
The resulting {\heulingo} solver is a tool for heuristically-driven
answer set optimization.
All source code including {\heulingo}
and benchmark problems is available
from \url{https://github.com/banbaralab/kr2024}.

The LNPS heuristic can be further
extended to \emph{adaptive LNPS} in which
a potentially more effective combination of
destruction and prioritization is selected
in each iteration.
The {\heulingo} approach can be applied to a wide range of
optimization problems.
In particular,
ASP-based LNPS for timetabling can be promising
since ASP has been shown to be highly effective for
curriculum-based course timetabling~\cite{basotainsc13a,bainkaokscsotawa18a}.
For this, in our preliminary experiments,
{\heulingo} succeeded in finding improved bounds for 8 instances
in the most difficult UD5 formulation,
compared with the best known bounds obtained by more
dedicated metaheuristics.
From a broader perspective,
integrating LNPS into MaxSAT and PB optimization would be interesting.
We will investigate these possibilities, and 
our results will be applied to solving real-world applications.

\bibliographystyle{kr}
\bibliography{krr,procs,local}

\end{document}